\documentclass[letterpaper, 10 pt, conference]{ieeeconf}
\usepackage{cite}
\usepackage{amsmath,amssymb,amsfonts} 
\usepackage{graphicx}
\usepackage{textcomp} 
\usepackage{multirow}
\usepackage{gensymb}
\usepackage{xcolor} 
\usepackage[ruled,vlined]{algorithm2e}
\usepackage{gensymb}
\usepackage{textcomp} 
\usepackage{xcolor} 
\usepackage{color}
\usepackage{multirow}
\usepackage{graphicx}
\usepackage{subfigure}
\usepackage{verbatim}
\usepackage[mathscr]{eucal}
\usepackage{lipsum,mwe,cuted}
\usepackage{float}
\usepackage{caption}
\usepackage{dblfloatfix}
\usepackage{blindtext}
\usepackage{bm}
\usepackage{url}

\usepackage[misc]{ifsym}

\usepackage[hang]{footmisc}

\IEEEoverridecommandlockouts                              

\makeatletter

\makeatletter
\renewcommand{\maketag@@@}[1]{\hbox{\m@th\normalsize\normalfont#1}}%

\setbox0\hbox{$\xdef\scriptratio{\strip@pt\dimexpr
    \numexpr(\sf@size*65536)/\f@size sp}$}

\newcommand{\scriptveryshortarrow}[1][3pt]{{%
    \vcenter{\hbox{\rule[\scriptratio\dimexpr-.2pt\relax]
               {\scriptratio\dimexpr#1\relax}{\scriptratio\dimexpr.4pt\relax}}}%
   \mkern-4mu\hbox{\let\f@size\sf@size\usefont{U}{lasy}{m}{n}\symbol{41}}}}

\makeatother

\title{\LARGE \bf
CROON: Automatic Multi-LiDAR Calibration and Refinement Method in Road Scene
}
\author{Pengjin Wei$^{*a}$, Guohang Yan$^{*b}$, Yikang Li$^b$, Kun Fang$^a$, Xinyu Cai$^b$, Jie Yang$^{{\dagger}a}$, Wei Liu$^{{\dagger}a}$
\thanks{$^{*}$ Equally contribution.}
\thanks{$^{\dagger}$ Corresponding author.}
\thanks{$^{a}$ P. Wei, K. Fang, W. Liu, J. Yang are with the Institute of Image Processing and Pattern Recognition; Department of Automation; Shanghai Jiao Tong University, China. \{pengjinwei, fanghenshao, liuwei.1989, jieyang\}@sjtu.edu.cn}
\thanks{$^{b}$ G. Yan, Y. Li, X. Cai are with the Autonomous Driving Group, Shanghai AI Laboratory, China. \{yanguohang, liyikang, caixinyu\}@pjlab.org.cn}
\thanks{$^{c}$ This research is partly supported by National Key R\&D Program of China(No.2019YFB1311503) and NSFC, China (No.61876107, U1803261, 61977046)}
}
\begin{document}
\maketitle
\begin{abstract}
Sensor-based environmental perception is a crucial part of the autonomous driving system. In order to get an excellent perception of the surrounding environment, an intelligent system would configure multiple LiDARs (3D Light Detection and Ranging) to cover the distant and near space of the car. The precision of perception relies on the quality of sensor calibration. This research aims at developing an accurate, automatic, and robust calibration strategy for multiple LiDAR systems in the general road scene. We thus propose \textbf{CROON} (automati\textbf{c} multi-LiDA\textbf{R} Calibrati\textbf{o}n and Refinement methOd in r\textbf{O}ad sce\textbf{N}e), a two-stage method including rough and refinement calibration. The first stage can calibrate the sensor from an arbitrary initial pose, and the second stage is able to precisely calibrate the sensor iteratively. Specifically, CROON utilize the nature characteristics of road scene so that it is independent and easy to apply in large-scale conditions. Experimental results on real-world and simulated data sets demonstrate the reliability and accuracy of our method. All the related data sets and codes are open-sourced on the Github website \url{https://github.com/OpenCalib/LiDAR2LiDAR}.
\end{abstract}

\section{INTRODUCTION}
\begin{figure}[ht]
\centering
\includegraphics[scale=0.15]{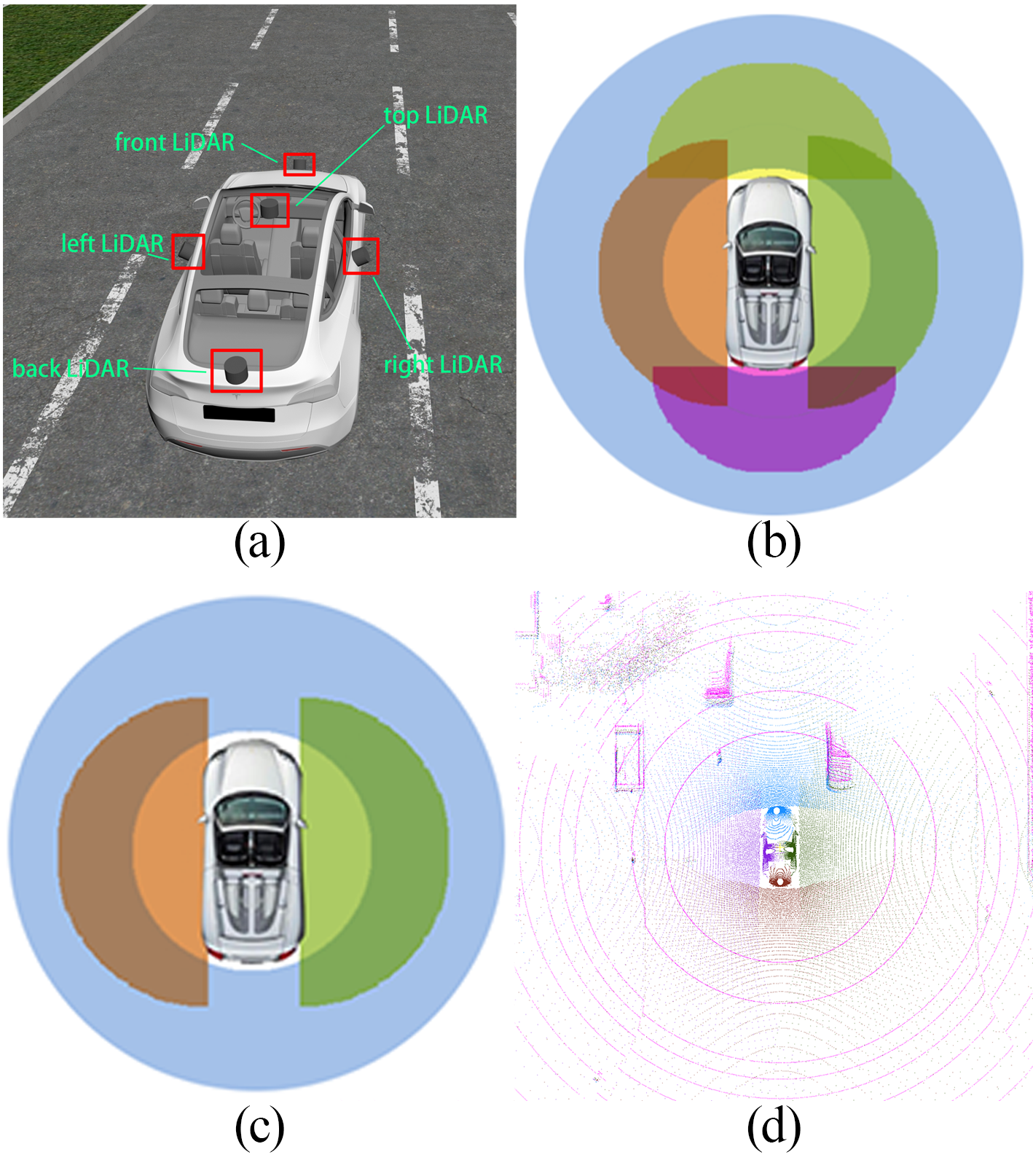}
\caption{(a)LiDARs configuration in the simulated engine, (b)the field of view of the top, front, back, left, right LiDAR, our unreal world data set is collected under this configuration, (c)the field of view of the top, left, right LiDAR, our real world data set is collected under this configuration, (d) a sample of aligned unreal world data.}
\label{LiDARs configuration}
\end{figure}

With the development of computing power, high precision sensors, and further research in visual perception, automatic driving naturally becomes the focus\cite{RN11,RN15,RN23}. 3D Light Detection and Ranging (LiDAR) has been the primary sensor due to its superior characteristics in three-dimensional range detection and point cloud density. In order to better perceive the surrounding environment, the scheme for a car with multi-LiDARs is necessary. Empirically, 5 LiDARs configured as shown in Fig. \ref{LiDARs configuration}(a) can cover the near and far field. In order to get a wider field of view and denser point cloud representation, it is worthwhile to accurately calculate the extrinsic parameters, i.e., the rotations and translations of the LiDARs, to transform the perceptions from multiple sensors into a unique one. 

The extrinsic parameters were calibrated manually at the very beginning. Now there are many sensor extrinsic calibration methods, and they can be divided into target-based ones and motion-based ones. Target-based methods need auxiliary things which have distinguished geometric feature. They solve the problem with the assistance of specific targets like retro-reflective landmarks\cite{RN14,RN19}, boxes\cite{RN17}, and a pair of special planes\cite{RN21}. On the contrary, Motion-based strategies estimate sensors pose by aligning an estimated trajectory and they deeply rely on vision and inertial sensors, e.g., cameras and inertial measurement unit (IMU), to enrich the constraints\cite{RN9,RN22}. Another type of target-less methods could alleviate the dependence on assistant items, but still require special application scenarios\cite{RN18,RN11}. They all facilitate the solution of the problem, but there are still several limitations among these methods. Target-based methods require to measure unique markers\cite{RN14,RN19,RN17,RN21}, and the measurement will inevitably introduce errors. Because of those initial preparations in advance, target-based methods can not be applied in large-scale cases. Motion-based methods are often used as rough calibration, and they can not get precise estimations due to equipment performance and trajectory\cite{RN11}. Although some target-less methods have been proposed, they need more strict initialization or environment so that can not be used efficiently. They still can not meet the requirements of robustness, accuracy, automation, and independence simultaneously.

To address these issues, we propose a novel automatic multi-LiDAR Calibration and Refinement methOd in rOad sceNe, CROON, a two-stage calibration method consisting of a rough calibration component followed by a refinement calibration component. The LiDAR from an arbitrary initial pose can be calibrated into a relatively correct status in the rough calibration. We develop an octree-based method to continue optimizing multiple LiDARs poses in the refinement calibration. 
This two-stage framework is target-less since CROON could utilize the natural characteristic of road scenes so that it has a low dependence on other assistant stuff and thus is much more applicable in various situations than those target-based and motion-based ones.
Further, even though with less targets, this two-stage framework and the associated carefully-designed optimization strategy could still guarantee the stronger robustness and higher accuracy than that of those target-less methods.
In short, in the premise of a low independence on the auxiliary targets, the proposed CROON holds no loss in the robustness and accuracy in the task of extrinsic parameters calibration, which has been verified by the extensive empirical results.

The contributions of this work are as follows: 
\begin{enumerate}
\item The proposed method is an automatic and target-less calibration method of multiple LiDARs in road scenes.
\item We introduce a rough calibration approach to constrain the LiDARs from large deviations and an octree-based refinement method to optimize the LiDARs pose after classic iterative closest point (ICP).
\item The proposed method shows promising performance on our simulated and real-world data sets; meanwhile, the related data sets and codes have been open-sourced to benefit the community.
\end{enumerate}

\section{RELATED WORK}

Extrinsic calibration aims to determine the rigid transformations (i.e., 3D rotation and translation) of the sensor coordinate frame to the other sensors or reference frames \cite{RN41}, The methods of extrinsic calibration for sensor systems with LiDARs can be divided into two categories: (1) motion-based methods that estimate relative poses of sensors by aligning an estimated trajectory from independent sensors or fused sensor information, and (2) target-based/appearance-based methods that need specific targets which have distinguished geometric feature such as a checkerboard.

The motion-based approaches are known as hand-eye calibration in simultaneous localization and mapping (SLAM). The hand-eye calibration was firstly proposed to solve mapping sensor-centered measurements into the robot workspace frame, which allows the robot to precisely move the sensor\cite{RN39,RN40,RN42}. The classic formulation of the hand-eye calibration is $AX = XB $, where $A$ and $B$ represent the motions of two sensors, respectively, and $X$ refers to the relationship between the two sensors. Heng et al.\cite{RN36} proposed a versatile method to estimate the intrinsic and extrinsic of multi-cameras with the aid of chessboard and odometry. Taylor et al.\cite{RN22, RN9}introduced a multi-modal sensors system calibration method that requires less initial information. Qin et al.\cite{RN35} proposed a motion-based methods to estimate the online temporal offset between the camera and IMU. They also proposed to estimate the calibration of camera-IMU transformation by visual-inertial navigation systems in an online manner\cite{RN34}. The research work in \cite{RN10} aligned the camera and lidar by using the sensor fusion odometry method. In addition, Jiao et al.\cite{RN11} proposed a fusion method that included motion-based rough calibration and target-based refinement. It is worth noting that motion-based methods heavily depend on the results of the vision sensors and can not deal with the motion drift well. 

Target-based approaches recover the spatial offset for each sensor and stitch all the data together. Therefore, the object must be observable to sensors, and correspondence points should exist. Gao et al.\cite{RN14} estimated extrinsic parameters of dual LiDARs using assistant retro-reflective landmark on poles which could provide distinguished reflect signal. Xie et al.\cite{RN13} demonstrated a solution to calibrate multi-model sensors by using Apriltags in the automatic and standard procedure for industrial production. Similarly, Liao et al.\cite{RN28} proposed a toolkit by replacing Apriltags with the polygon. Kim et al.\cite{RN15} proposed an algorithm using the reflective conic target and calculating the relative pose by using ICP. Those methods\cite{RN13,RN15} complete final calibration by using the parameters of special targets known in advance, so it is essential to estimate the exact measurement. Zhou et al.\cite{RN25} demonstrated a method to facilitate the calibration by establishing line and plane relationship in chessboard. Z. Pusztai\cite{RN17} introduced a method that mainly required the perpendicular characteristic of boxes for accurate calibration in the camera-LiDAR system. Choi et al.\cite{RN21} designed an experiment to estimate the extrinsic parameters of two single-line laser-line sensors by using a pair of orthogonal planes. Similarly, three linearly independent planar surfaces were used in automatic LiDAR calibration work in\cite{RN18}.

The real automatic and target-less calibration methods attract more and more attention. He et al.\cite{RN29,RN30} extracted point, line, and plane features from 2D LIDARs in nature scenes and calibrated each sensor to the frame on a moving platform by matching the multi-type features. Automatic extrinsic calibration of 2D and 3D LIDARs method was put forward in \cite{RN32}. Pandey et al.\cite{RN33} demonstrated the mutual information based algorithm could benefit the calibration of a 3D laser scanner and an optical camera.

\section{METHODOLOGY}

 In our experiment, we have five LiDARs. They can be divided into two classes of the master and slaves. The top LiDAR is the master LiDAR $PC_{m}$ and the rest (front, back, left, right) are slave LiDARs $PC_{s}(PC_{f}, PC_{b}, PC_{l}, PC_{r})$. Our goal is calibrating LiDARs by estimating the extrinsic parameters including rotation $\bm{R}$ and translation $\bm{T}$ and finally fusing all LiDARs data into a single one as demonstrated in Fig.\ref{LiDARs configuration}(d). $\bm{R}$ and $\bm{T}$ can be subdivided into $angle_{pitch},angle_{roll},angle_{yaw}$ and $x,y,x$, that represent the rotation angle and translation in three dimensions, respectively. The problem can be defined as follow:

\begin{equation}\label{origin cost function}
\begin{aligned}
\bm{R}^*,\bm{T}^*\ =\ \underset{\bm{R} ,\bm{T}}{\arg\min}\sum\limits _{(p_{m_i},p_{s_i})\in C} \lvert\lvert \bm{R}\cdot p_{m_i} + \bm{t} - p_{s_i} \rvert\rvert_2^2
\end{aligned} 
\end{equation}

\noindent where $\lvert\lvert\cdot\rvert\rvert_2 $ indicates the $l_{2}$-norm of a vector. $p_{m_i}$ from $PC_m$ and $p_{s_i}$ from $PC_s$ are correspondences.

\begin{figure*}[t]
\centering
\includegraphics[scale=0.2]{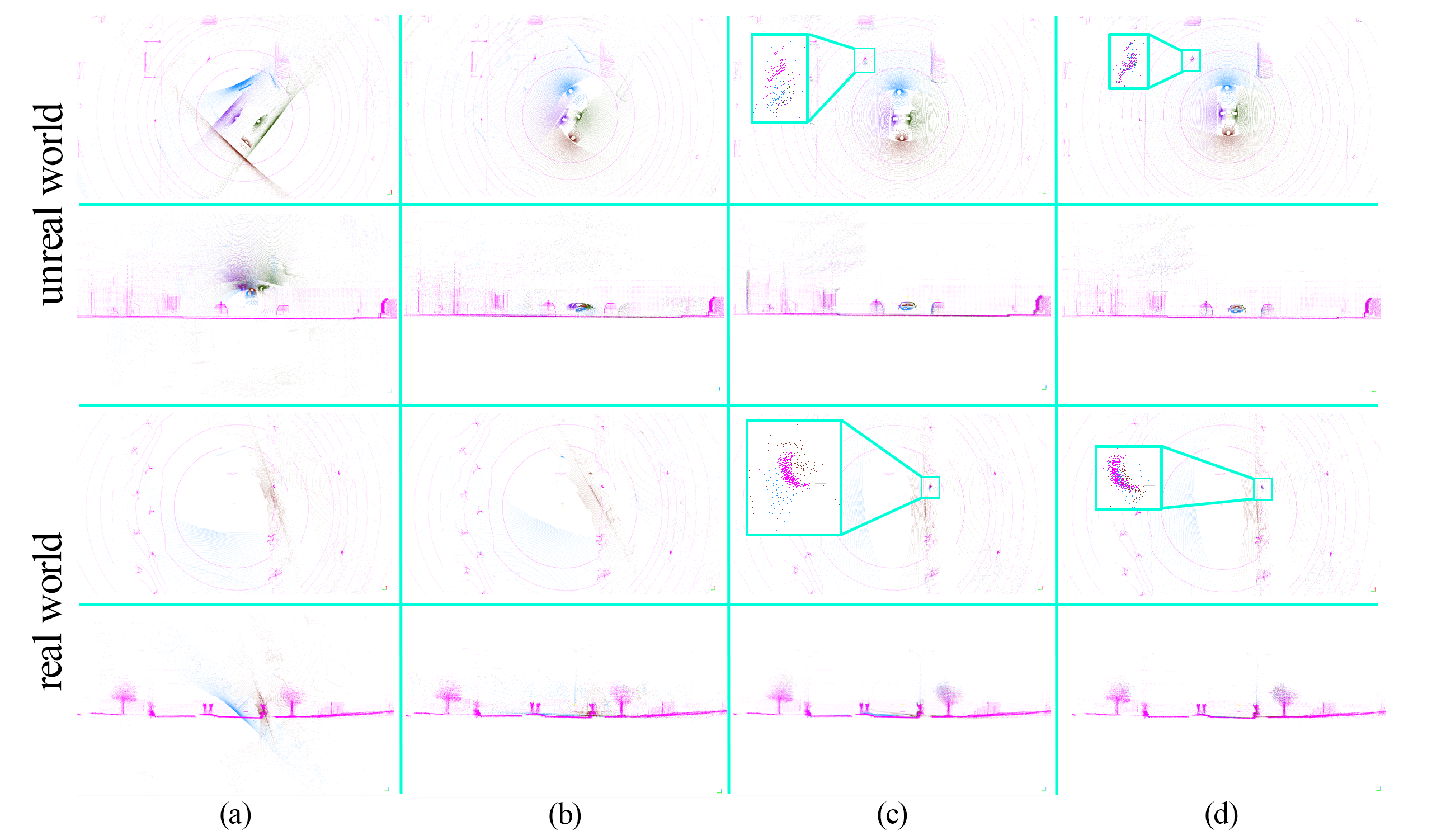}
\caption{The different periods of our method in real/unreal world data set. The first two rows represent the top-view and left-view of unreal world data. The last two rows represent the top-view and left-view of real world data. (a) column shows the initial pose, (b) column is collected when the ground planes of point clouds are calibrated, (c) column represents the results after rough calibration, and (d) column represents the renderings of refinement calibration.}
\label{aligned data}
\end{figure*}

It is not tractable to optimize a $ 6-DOF $ problem from the beginning because the range of the parameters is too large. So in the first section, the solution which could narrow the range quickly and robustly will be proposed. 

\subsection{rough calibration}
The method we proposed is applied to road scene which widely exist. On the road, the LiDAR can easily sample a large amount of ground plane information, therefore, ground plane always can be registrated correctly. It means that $angle_{pitch}, angle_{roll}, and z$ are solved. So the first step of our algorithm is rough registration by taking advantage of the characteristic. \cite{Zaiter} proposed similar approach utilizing the ground plane feature but we adjust the registration process. In our method, we first calibrate the $angle_{pitch}, angle_{roll}, and z$ with ground plane and determine the scope of $angle_{yaw}$ quickly by scanning the definition domain.

Suppose the maximal plane which contains the most points is considered as the ground plane $GP:\{a,b,c,d\}$: 
\begin{equation}\label{GP extration}
\begin{aligned}
GP:\{a,b,c,d\} = \arg\underset{\lvert plane \rvert}{\max} \lvert {ax_i + by_i + cz_i + d \rvert \leq \epsilon} 
\end{aligned} 
\end{equation}
where $(x_i,y_i,z_i) \in plane$ and $plane \subset  PC$ and $\epsilon$ means the threshold of the plane thickness.
The ground plane is used to align the slave LiDAR ground planes $GP_{s}(GP_{f}, GP_{b}, GP_{l}, GP_{r})$ to the master LiDAR ground plane $GP_{m}$:
\begin{equation}\label{GP registration axis}
\begin{aligned}
\vec{n} = \overrightarrow{GP_m} \times \overrightarrow{GP_s}
\end{aligned} 
\end{equation}

\begin{equation}\label{GP registration angle}
\begin{aligned}
\theta = \overrightarrow{GP_m} \cdot \overrightarrow{GP_s}
\end{aligned} 
\end{equation}
where $\vec{n}$, $\theta$, $\overrightarrow{GP_m}$, $\overrightarrow{GP_s}$ represent the rotation axis, rotation angle, master LiDAR normal vector and slave LiDAR normal vector, respectively. The transformation matrix can be computed by Rodriguez formula.

It is worth noting that an extreme case can appear where the difference between the estimated pitch/roll and the actual pitch/roll is $\pm \pi$. So the method need to check whether most of the points of $PC_{s}$ are on the ground plane after the calibration. According to the point cloud without ground points, the direction of the ground plane normal vector can be confirmed quickly, and all $GP$ normal vectors should have the same direction, $\xrightarrow{}$ direction as $\overrightarrow{GP_m} = \overrightarrow{GP_s}$.


Through above measures, a rough estimation of $angle_{pitch}, angle_{roll}, z $ can be established. The next step is the calibration of $angle_{yaw}, x, y $. The cost function can be simplified from (\ref{origin cost function}):
\begin{equation}\label{simplified cost function}
\begin{aligned}
&angle^{*}_{yaw} ,\ x^*,\ y^*\ = 
\\ \ &\underset{yaw,x,y}{\arg\min}\sum\limits _{(p_{m_i},p_{s_i})\in C} \lvert\lvert \bm{R_{yaw}}\cdot p_{m_i} + \bm{x} +\bm{y} - p_{s_i} \rvert\rvert_2^2
\end{aligned} 
\end{equation}
Here, $R_{yaw}$ indicates the matrix which only include $angle_{yaw}$ information, $x$ and $y$ are the deviation in $X-axis$ and $Y-axis$. $p_{m_i}$ and $p_{s_i}$ are correspondences which has correct $angle_{pitch}$, $angle_{roll}$ and $z$. The number of arguments decreases from 6 to 3. More importantly, the ground points could be ignored. There are three obvious advantages of the method described above: a) the computational complexity is reduced significantly because the proportion of ground points is very high; b) the points with more explicit characteristics are extracted because the relatively ordinary ground points have been discarded; c) points that should have a more significant impact on the cost function are sampled because the error of distant points can better reflect the effect in practice. 

In real application, we approximately solve Eq.\ref{simplified cost function} by firstly finding the optimal $angle_{yaw}$ and then the optimal ones of $x$ and $y$. This is because the pose error caused by rotation effects more on the loss function than that cause by the translation. We choose to scan the range of $angle_{yaw}$ by using binary search.

\subsection{refinement calibration}
After the rough registration described in the last subsection, we can further refine the relative right pose for each LiDAR. In this section, we will continue improving the calibration accuracy by iterative closest points with normal (ICPN)\cite{serafin2014using,serafin2015nicp,serafin2017using} and octree-based optimization.

We first use a variant of ICP. The origin ICP is to find the optimal transform between two point clouds that minimizes the loss function $ \sum\limits _{n=1}^{N} \lvert\lvert PC_{s} -PC_{m} \rvert\rvert $. The method requires strict initialization, and it is easy to trap in the locally optimal value. We thus adopt the ICPN which is a variant of ICP and can achieve better performance. We suppose that due to the sparsity of the point cloud, the point cloud feature is not explicit and hard to extract. The ICPN enriches the point feature by containing the normal of each point. A point normal includes position information of the point, and more importantly, it could provide the neighbor points information. These together make the ICPN expands the receptive field of every point and better utilizes the local information for calibration. In our implementation, the normal of each point is calculated from the nearest 40 neighboring points by PCA.
\begin{figure}[ht]
\centering
\includegraphics[scale=0.2]{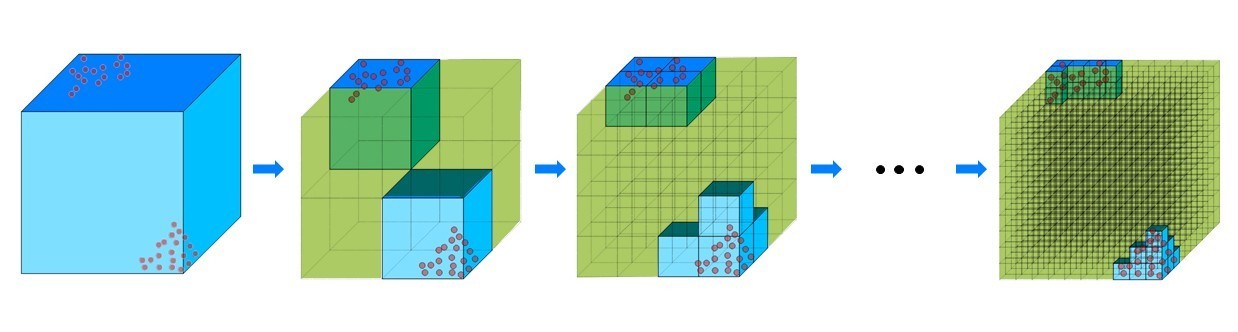}
\caption{Octree-based method. Mark the cube with points in blue and mark the cube without a point in green, cutting the cube iteratively and the volume of blue/green cubes can be measured the quality of calibration.}
\label{octree}
\end{figure}

\begin{figure*}[ht]
\centering
\includegraphics[scale=0.2]{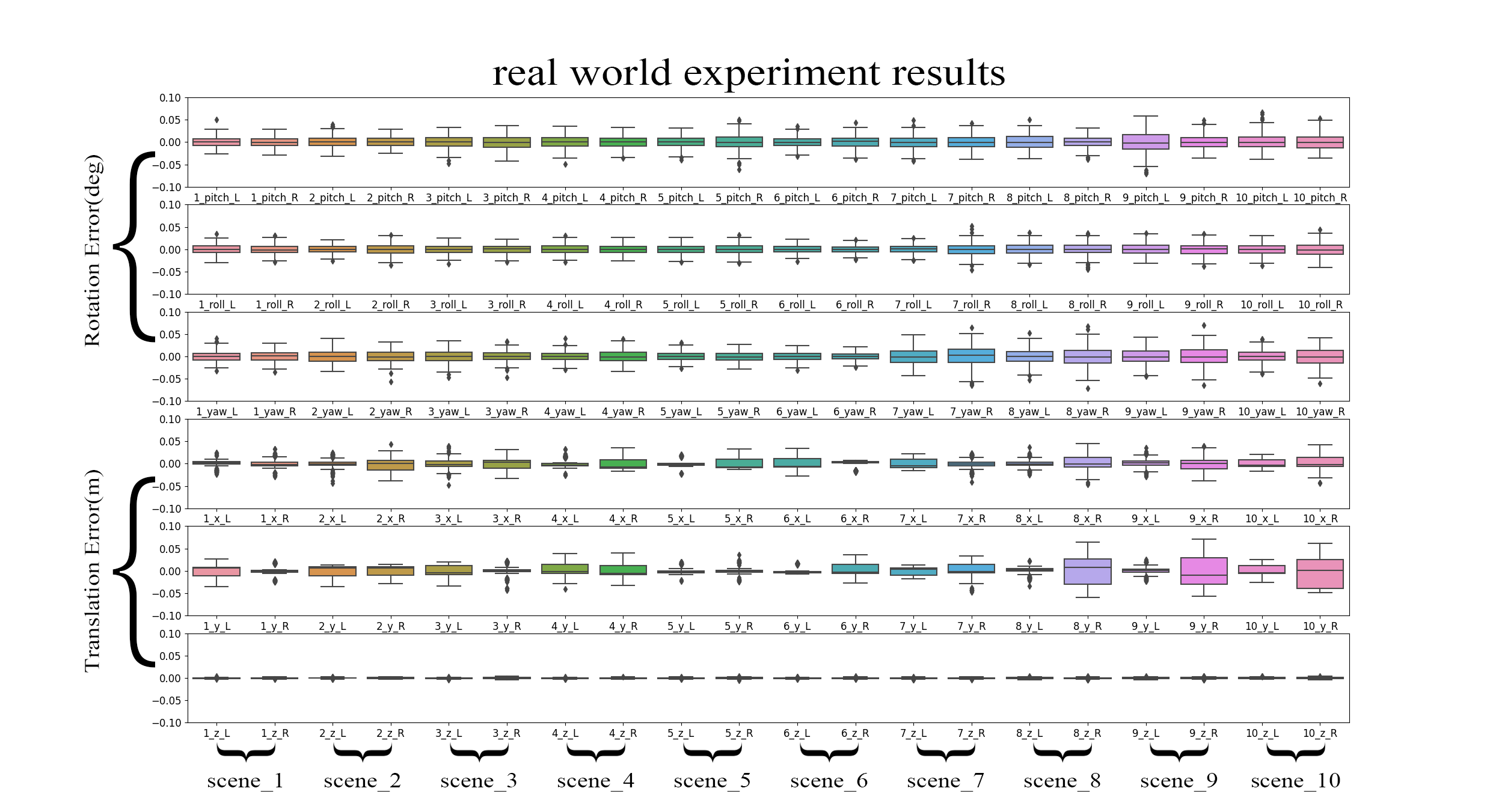}
\caption{Experiment results of ten real road scenes. In each scene, we collect 250 sets of data. We repeat the calibration 250 times and calculate the mean and standard deviation of the 6-DOF results.}
\label{plotbox of real scene}
\end{figure*}

Furthermore, we continue minimizing the pose error with the octree-based optimization as illustrated in Fig.\ref{octree}. At the beginning, there are two point clouds $PC$ wrapped in a cube $_{o}C$. We then utilize the octree-based method to equally cut the cube into eight smaller cubes. 
\begin{equation}\label{cut the cube}
\begin{aligned}
 _{p}C \stackrel{cut }{\Longrightarrow}  \{_{c}C_1, _{c}C_2,\cdots, _{c}C_7, _{c}C_8\}
\end{aligned} 
\end{equation}
where $_{p}C$ represents the parent cube and $_{c}C_{i}$ represent the child cube of $_{p}C$. The cutting procedure is iteratively repeated to get more and smaller cubes. We mark the cubes with points as blue ones and the cubes without a point as the green ones, as shown in Fig.\ref{octree}. They are further denoted as $C^b$ and $C^g$. The volume of $_{o}C$ can be expressed as follow:
\begin{equation}\label{calculate the space}
\begin{aligned}
V_{_{o}C} = \sum\limits _{i=1}^{N} V_{C^b_i} + \sum\limits _{j=1}^{M} V_{C^g_j}
\end{aligned} 
\end{equation}
where $N$ and $M$ refer to the number of $C^b$ and $C^g$. When the side length of the small cube is short enough, We can approximate that the space volume occupied by the point cloud is the volume of the blue cubes. When two point clouds are aligned accurately, the space volume occupied by point clouds reaches the minimum, and the volume of blue cubes reaches the minimum at the same time. So the problem can be converted to:
\begin{equation}\label{octee-based optimal}
\begin{aligned}
\bm{V_{occupy\_space}}\ =\ \underset{}{\min} \sum\limits _{j=1}^{M} V_{C^b_j}
\end{aligned} 
\end{equation}
Considering that the current pose is close to the correct value, we continue optimizing the formula above by scanning the domain of arguments.

\section{EXPERIMENTS}
In this section, we will apply our method in two different style data sets. The real data set is collected from our driverless vehicle platform. The driveless vehicle platform configures three LiDARs in the top, left, and right of the vehicle. The three LiDARs are high-precision and have a large view field, 10 refresh rate, and well-calibrated intrinsics. Another data set is collected from Carla simulated engin\cite{Dosovitskiy17}. The driverless vehicle in the unreal engine has more LiDARs in the front and back. To analyze the accuracy, robustness, and efficiency of the proposed calibration method, we test on a number of different road conditions. Experiment results show that our method achieves better performance than those state-of-the-art approaches in terms of accuracy and robustness.

\begin{table*}[b]
\centering
\setlength{\tabcolsep}{5mm}
\renewcommand{\arraystretch}{1.2}
\caption{Quantitative results on simulated data set.}
\begin{tabular}{cccccccc}
\hline
\multicolumn{2}{c}{\multirow{2}{*}{\textbf{LiDAR position}}} & \multicolumn{3}{c}{\textbf{Rotation Error{[}deg{]}}} & \multicolumn{3}{c}{\textbf{Translation Error{[}m{]}}} \\ \cline{3-8} 
\multicolumn{2}{c}{}                                         & \textbf{pitch}    & \textbf{roll}   & \textbf{yaw}   & \textbf{x}       & \textbf{y}       & \textbf{z}      \\ \hline
\multirow{2}{*}{front LiDAR}         & \textbf{mean}         & -0.0111        & -0.0124      & 0.0049     & 0.0028       & -0.0005     & -0.0013     \\ \cline{2-8} 
                                     & \textbf{std}          & 0.0407         & 0.0310       & 0.0875      & 0.0363        & 0.0075       & 0.0030      \\ \hline
\multirow{2}{*}{back LiDAR}          & \textbf{mean}         & -0.0108        & 0.0163       & -0.0234     & -0.0163       & -0.0007     & -0.0010    \\ \cline{2-8} 
                                     & \textbf{std}          & 0.0266         & 0.0363       & 0.1451       & 0.1070         & 0.0149        & 0.0037      \\ \hline
\multirow{2}{*}{left LiDAR}          & \textbf{mean}         & -0.0220       & -0.00004    & 0.0029     & -0.0033      & -0.0005     & -0.0014     \\ \cline{2-8} 
                                     & \textbf{std}          & 0.0290         & 0.0106       & 0.0760      & 0.0460        & 0.0114         & 0.0022      \\ \hline
\multirow{2}{*}{right LiDAR}         & \textbf{mean}         & 0.0105        & 0.0087      & 0.0133      & -0.0003     & 0.0020       & 0.0011      \\ \cline{2-8} 
                                     & \textbf{std}          & 0.0291        & 0.0293       & 0.0747      & 0.0446        & 0.0075       & 0.0075      \\ \hline
\end{tabular}
\label{unreal world statistic results}
\end{table*}

Because there is no ground truth data for the real data set, we thus only qualitatively evaluate the results. Compared with real data, simulated data has ground truth, and the method can be tested completely through quantitative and qualitative analysis.

\subsection{Realistic Experiment}
In this section, we first collect real-world point cloud data in several different road scenes in our city. It should be noted that the LiDARs have an initial configuration angle and position offset. We add a random deviation to LiDARs, including $\pm 45\degree$ in $pitch, roll, yaw$, and $\pm 10cm $ in $x, y, z$. After adding the artificial deviation and continuous measurement in one scene, we can evaluate its consistency and accuracy. Furthermore, the results of different scenes can show its robustness and stability.


\subsubsection{Qualitative Results}
Our method consists of two stages, rough calibration and refinement calibration. The last two rows of Fig.\ref{aligned data} show the different period point clouds in real scene. The third row of Fig.\ref{aligned data} is the top view, and the fourth row of Fig.\ref{aligned data} is the left view. Columns of (a)-(d) in Fig.\ref{aligned data} represent the initial pose, ground plane calibration, rough calibration, and refine calibration, respectively. Point clouds from three different LIDARs have a large initial deviation in column (a). After ground plane calibration, three point clouds calibrate their non-ground points in rough calibration and refine calibration in column (c) and (d) where the pose of the three point clouds are quite accurate.

\begin{table*}[]
\centering
\setlength{\tabcolsep}{5mm}
\caption{Compare results with other two targetless method, CROON achieves good results in 6 degrees of freedom.}{
\begin{tabular}{ccllcccc}
\hline
\multirow{2}{*}{Method}                                                            & \multirow{2}{*}{scene} & \multicolumn{3}{c}{Rotation{[}deg{]}}                                      & \multicolumn{3}{c}{Translation{[}m{]}}                                                                   \\ \cline{3-8} 
                                                                                  &                        & pitch            & roll             & yaw                                  & x                           & y                                   & z                                    \\ \hline
\multirow{3}{*}{\begin{tabular}[c]{@{}c@{}}Single \\ Planar \\ Board\cite{RN20}\end{tabular}} & config1                & -0.0264          & -1.0449          & -0.3068                              & 0.0094                      & -0.0098                             & 0.0314                               \\ \cline{2-8} 
                                                                                  & config2                & 0.1587           & -0.3132          & -0.7868                              & \textbf{0.0026}             & \textbf{0.0029}                     & \textbf{-0.0105}                     \\ \cline{2-8} 
                                                                                  & config3                & 1.023            & -0.0902          & 0.1441                               & \textbf{-0.0011}            & -0.0033                             & -0.0162                              \\ \hline
\multirow{2}{*}{\begin{tabular}[c]{@{}c@{}}Planar \\ Surfaces\cite{RN18}\end{tabular}}        & config1                & \textbf{-0.0097} & \textbf{-0.0084} & -0.0149                              & -0.0407                     & 0.0358                              & 0.0416                               \\ \cline{2-8} 
                                                                                  & config2                & 0.0172           & 0.0165           & \textbf{-0.0017}                     & 0.0541                      & 0.0123                              & 0.0632                               \\ \hline
CROON                                                                              & all                    & \textbf{-0.0083} & \textbf{0.0031}  & \multicolumn{1}{l}{\textbf{-0.0006}} & \multicolumn{1}{l}{-0.0043} & \multicolumn{1}{l}{\textbf{0.0001}} & \multicolumn{1}{l}{\textbf{-0.0006}} \\ \hline
\end{tabular}
}
\label{compare results}
\end{table*}
\subsubsection{Calibration Consistency Evaluation}

We test our method in ten different scenes where 250 groups of different initial values are used in each scene. In Fig.\ref{plotbox of real scene}, the $x$-axis represents the different road scenes, and the $y$-axis represents the mean and standard deviation of 6 degrees of freedom. In each scene, the means of different rounds of measurement are close to each other, which demonstrates the consistency and accuracy of our method. More essentially, all the standard deviation in different scenes are close to 0, which shows the robustness and stability of our method. It is worth mention that our method can get excellent results in $z$ translation calibration due to utilizing the characteristic of road scenes.





\subsection{Simulated Experiment}
We can get accurate ground truth data in the simulated engine by controlling the other influence factors compared with real-world data. In the simulated experiment, we first collect simulated point cloud data in different road scenes in the Carla engine. We collect the data while the virtual car automatically running, and we can directly consider that all LiDARs are time synchronization. The unreal world data set totally includes 1899 frames of point clouds. The simulated experiment results are mainly used for quantitative analysis to demonstrate the robustness, stability, and accuracy of our method again.

\subsubsection{Quantitative Experiments}
\begin{figure}[ht]
\centering
\includegraphics[scale=0.12]{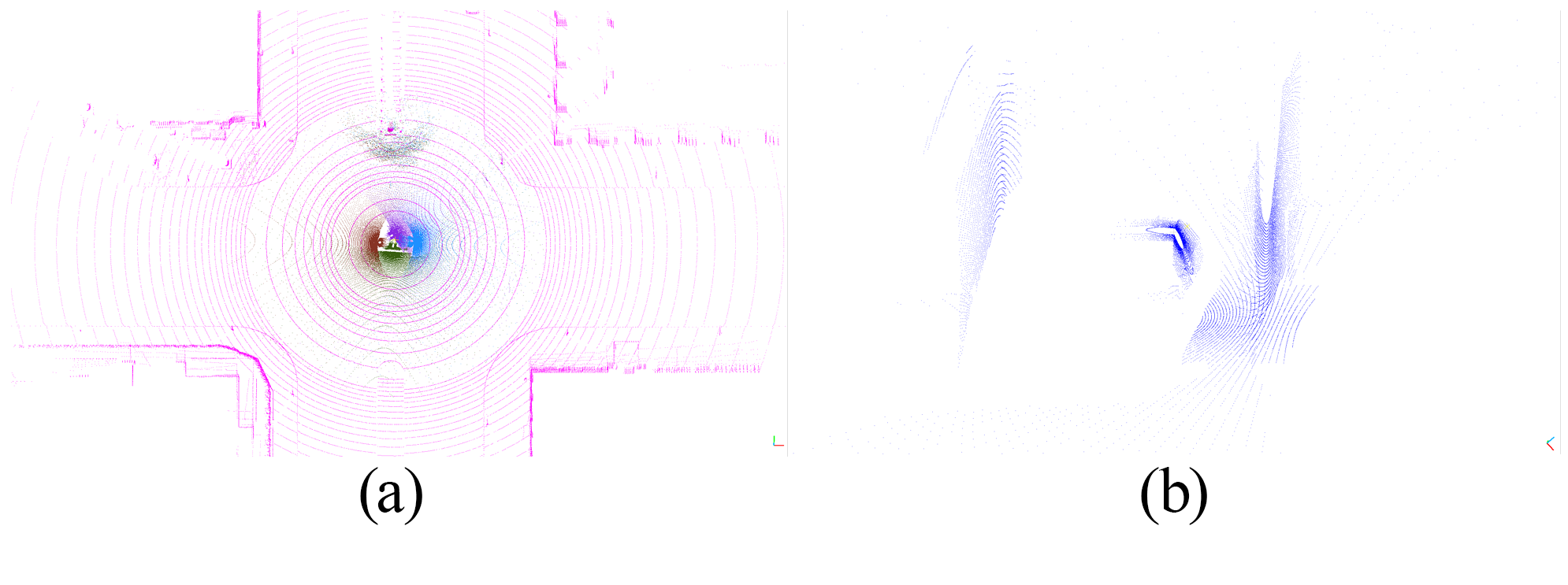}
\caption{Bad case. (a) All around is the same in the scene, (b) There is a truck in the right of the car to block the right lidar.}
\label{simulated data set bad case}
\end{figure}

Similar to the experiments of the real-world data set, We randomly initialize the parameters and then calculate the means and standard deviations of all results. our method can calibrate more than 1798(94.7\%) road scenes successfully and Table.\ref{unreal world statistic results} shows the quantitative results. There are some failure cases in Fig.\ref{simulated data set bad case}. Our method calibrates the LiDARs with single frame point cloud. So when the road scene is very empty and simple, or extremely narrow, our method will fail, Because in this two situation, the point clouds from LiDARs are all useless signals.

\subsubsection{Comparison Experiments}
We compare our methods with \cite{RN18} and \cite{RN20}, which all perform automatic calibration methods and use prior information as little as possible. Table.\ref{compare results} shows the quantitative comparison of three methods. Our method can get the best results in $angle_{pitch}, angle_{roll}, angle_{yaw}, y, z$. It should be pointed out that our method is estimated under thousands of groups of significant random initial extrinsic error and different road scenes.

\section{CONCLUSIONS}
In this paper, we propose CROON,a LiDAR-to-LiDAR automatic extrinsic parameters calibration method in road scene to find a set of stunning precision transformations between front, back, left, right LiDARs and top LiDAR. The method is a rough-to-fine framework to calibrate from an arbitrary initial pose accurately. Meanwhile, because of more geometric constraints from the raw data and the characteristic of the road scene, our method could fast calculate the result independently. More essentially, all the source code and data, including real scenes and simulated scenes in this paper, are available to benefit the community.

\bibliographystyle{IEEEtran}
\bibliography{egbib}

\end{document}